\documentclass[sigconf]{acmart}

\def\BibTeX{{\rm B\kern-.05em{\sc i\kern-.025em b}\kern-.08emT\kern-.1667em\lower.7ex\hbox{E}\kern-.125emX}}

\usepackage{lineno,hyperref}
\modulolinenumbers[5]

\usepackage{amsmath,amsfonts}
\usepackage{algorithmic}
\usepackage{graphicx}
\usepackage{textcomp}
\usepackage{xcolor}

\usepackage{enumitem}

\usepackage{booktabs}
\usepackage{multirow}
\usepackage{dsfont}

\usepackage{graphicx}
\usepackage{caption}
\usepackage{subcaption}

\usepackage{amsthm}

\usepackage[linesnumbered,ruled]{algorithm2e}
\usepackage{ifthen}
\usepackage{amsmath}
\usepackage{amsfonts}
\usepackage{booktabs}
\newlength{\Oldarrayrulewidth}

\usepackage{color, colortbl}
\definecolor{Gray}{gray}{0.9}

\usepackage{changepage}

\usepackage{setspace}

\begin{document}

\title[Driving Style Representation by D-CRNN]{Driving Style Representation in Convolutional Recurrent Neural Network Model of Driver Identification}
\titlenote{All rights reserved to the authors, and The Ohio State University (2021).}

\author{Sobhan Moosavi}
\email{moosavi.3@osu.edu}
\affiliation{%
  \institution{Ohio State University and Lyft Inc.}
  \city{San Francisco}
  \state{CA}
}

\author{Pravar D. Mahajan}
\email{pravar.d.mahajan@gmail.com }
\affiliation{%
  \institution{Ohio State University and Google Inc.}
  \city{San Jose}
  \state{CA}
}

\author{Srinivasan Parthasarathy}
\email{srini@cse.ohio-state.edu}
\affiliation{%
  \institution{Ohio State University}
  \city{Columbus}
  \state{OH}
}

\author{Colleen Saunders-Chukwu}
\email{saundec9@nationwide.com}
\affiliation{%
  \institution{Nationwide Mutual Insurance Co.}
  \city{Columbus}
  \state{OH}
}

\author{Rajiv Ramnath}
\email{ramnath@cse.ohio-state.edu}
\affiliation{%
  \institution{Ohio State University}
  \city{Columbus}
  \state{OH}
}

\renewcommand{\shortauthors}{Moosavi et al.}

\begin{abstract}
Identifying driving styles is the task of analyzing the behavior of drivers in order to capture variations that will serve to discriminate different drivers from each other. This task has become a prerequisite for a variety of applications, including usage-based insurance, driver coaching, driver action prediction, and even in designing autonomous vehicles; because driving style encodes essential information needed by these applications. In this paper, we present a deep-neural-network architecture, we term {\em D-CRNN}, for building high-fidelity representations for driving style, that combine the power of convolutional neural networks (CNN) and recurrent neural networks (RNN). Using CNN, we capture semantic patterns of driver behavior from trajectories (such as a turn or a braking event). We then find temporal dependencies between these semantic patterns using RNN to encode driving style. We demonstrate the effectiveness of these techniques for {\em driver identification} by learning driving style through extensive experiments conducted on several large, real-world datasets, and comparing the results with the state-of-the-art deep-learning and non-deep-learning solutions. These experiments also demonstrate a useful example of bias removal, by presenting how we preprocess the input data by sampling {\em dissimilar} trajectories for each driver to prevent spatial memorization. Finally, this paper presents an analysis of the contribution of different attributes for driver identification; we find that {\em engine RPM}, {\em Speed}, and {\em Acceleration} are the best combination of features.
\end{abstract}


\maketitle


\section{Introduction}
\label{sec:intro}
Analysis of telematics data with the goal of learning {\it driving style} for an individual driver has become feasible thanks to the availability of different means to collect and store large amounts of driving data. Learning driving style is the task of capturing {\em variations} in driving behavior for different drivers, whereby using such variations, one can discriminate different drivers from each other based on {\it how} they drive. The problem of learning driving style is specifically important for driver risk prediction used by insurance companies \cite{wang2018you,chen2019graphical,moosavi2017characterizing}. The shift towards more personalized insurance products has created a need for understanding driving style at an individual level. Additionally, learning driving style is also useful for driver coaching in order to help drivers to improve their skills \cite{stanton2007changing}; driver action prediction to prevent dangerous events such as accidents \cite{olabiyi2017driver,jain2016brain4cars}; and in designing autonomous vehicles that mimic actual drivers \cite{kuderer2015learning}. 


Several approaches have been recently proposed for learning driving style 
\cite{miyajima2007driver,phumphuang2015driver,enev2016automobile,dong2016characterizing,fugiglando2017characterizing,fung2017driver,ezzini2018behind,hallac2016driver,chowdhury2018investigations,wang2017driver,el2019improving}. Learning driving style, to be formalized as a pattern recognition task, is about constructing a model that seeks to predict the identity of {\em driver} for an input {\em trajectory}, using ``style and behavioral variation information'' extracted from trajectory data. In other words, this task is somewhat equivalent to {\em trajectory classification}, where labels are drivers' identities. However, unlike studies such as \cite{kieu2018distinguishing}, we aim in using driving behavior information instead of spatial information to label trajectories, which is a more complicated task and provides more benefits and insights for the aforementioned applications. 

Limitations to prior work include using small set of drivers and trajectories; no correction for bias in data (such as high spatial similarities between the trajectories for a given driver) which leads to utilizing spatial information instead of learning style; over-simplified models which do not fully extract and utilize driving behavior data; and the cost of data collection -- such as by using specially equipped vehicles to collect data. 
Mitigating the impact of spatial similarity is to {\it correctly} assess ability of different models to encode ``driving style information'', which is an important objective in this study. Suppose for drivers $d_1$ and $d_2$, $R_1$ and $R_2$ represent their set of prevalent distinct routes, which they take everyday. Given these sets, we can train a classifier to properly distinguish between trajectories taken from $d_1$ and $d_2$. This is because trajectories taken from $d_1$ (which comprise routes in set $R_1$) are significantly different from those taken from $d_2$ (which comprise routes in set $R_2$). Such a classifier would largely utilize spatial similarity information, which is not an objective when we seek to derive and utilize driving style information. 

In this paper, we present a new framework that addresses these limitations. Our data is extensive and collected from the {\em CAN-bus}\footnote{Controller Area Network (or CAN-bus) is a communication mechanism to transfer a variety of information related to operation of a vehicle.}, that includes a wide range of data elements regarding the operation of a vehicle, as well as other sensory data including GPS, accelerometer, and magnetometer. We propose several data sampling strategies to limit the {\em spatial similarity} between trajectories and obtain a diverse set of trajectories for each driver. We also study the importance of different attributes to explore driving style, and show the greater discriminative power of a selected set of attributes coming from CAN-bus in comparison to the GPS and other sensory data. To perform driver identification via encoding driving style, we present a deep-neural-network architecture, named {\em D-CRNN}, which combines several components, including convolutional neural network (CNN), recurrent neural network (RNN), and fully connected (FC) component. Using the CNN component, we derive semantic information about driving patterns (a turn, a braking event, etc.). Using the RNN component of the architecture, we capture temporal dependencies between different driving patterns to encode behavioral information. Finally, the FC uses the encoded behavioral information to build a model that properly predicts the driver's identity when given a trajectory as input. We show the effectiveness of our proposal in comparison to the state-of-the-art techniques, through extensive experiments on several large, real-world datasets provided to us via our industry collaborations. 

This paper makes the following contributions: 
\begin{itemize}[leftmargin=12pt]
    \item We present a deep-neural-network architecture to capture driving style information from telematics data. We show the superiority of our proposal in comparison to the state-of-the-art via extensive experiments based on real-world data. 
    
    \item We demonstrate a useful example of ``bias removal'', by presenting how we preprocess the input data by sampling dissimilar trajectories for each driver, in order to ensure our model is not simply using {\it spatial memorization} to discriminate drivers. 
    
    \item We explore the differential discriminative power of various attributes for characterizing the driving style. Identifying which attributes are most relevant to driver identification may serve to achieve application goals, such as driver coaching.
\end{itemize}

\noindent The rest of the paper is organized as follows: we formalize the problem in Section~\ref{sec:problem}, and review the related work in Section~\ref{sec:related}. The model is presented in Section~\ref{sec:architecture}. Details of dataset and results are discussed in Sections \ref{sec:dataset} and \ref{sec:results}, respectively, and we conclude in Section~\ref{sec:conclusion}. 
\section{Problem Statement}
\label{sec:problem}
Suppose we have a database $\mathcal{T} = \big \{T_1, T_2, \dots, T_N \big \}$ of $N$ trajectories, where each trajectory $T$ is a time-ordered sequence of data points $\langle p_1, p_2, \dots, p_{|T|}\rangle$. We represent each data point $p$ by a tuple $\big( t, Speed, Accel,$ $Rpm, Lat, Lng, Head, AclX, AclY, AclZ\big)$, where $t$ is the timestamp; $Speed$ shows the ground velocity of the vehicle (metric is $m/s$); $Accel$ is the acceleration or the rate of change of velocity (metric is $m/s^2$); $Rpm$ represents engine's revolutions per minute; $Lat$ and $Lng$ represent positional latitude and longitude data; $Head$ shows the heading (or bearing) of the vehicle which is a number between 0 and 359 degrees\footnote{Heading (or bearing) value 0 shows the \textit{north} and 180 shows the \textit{south} direction.}; and $AclX$, $AclY$, and $AclZ$ represent the data coming from the accelerometer sensor which show the acceleration of the vehicle toward different axes in a three dimensional (3D) coordinate system. Additionally, suppose that we have a look-up table which maps each trajectory $T \in \mathcal{T}$ to a driver id $d$, where $d \in \mathcal{D}$, and $\mathcal{D}$ is the set of all drivers. Given the preliminaries, we formalize our problem as follows:

\vspace{3pt}
\noindent\textbf{\normalsize Input}: A trajectory $T$. 

\vspace{3pt}
\noindent\textbf{\normalsize Model}: A predictive model $M$ to capture variations in driving behavior to derive driving style information. 

\vspace{3pt}
\noindent\textbf{\normalsize Goal}: Predict identity of driver for trajectory $T$ based on driving style information.

\vspace{3pt}
\noindent\textbf{\normalsize Optimization Objective}: Minimize prediction error. 


\section{Related Work}
\label{sec:related}
Driver identification by learning driving style from telemetry data has been the topic of much prior research work  \cite{miyajima2007driver,lopez2012driver,phumphuang2015driver,dong2016characterizing,ezzini2018behind,hallac2016driver,wang2017driver,chowdhury2018investigations,hallac2018drive2vec}.


Miyajima et al. \cite{miyajima2007driver} employed car following patterns and pedal operations (i.e., gas and break) as two important sources of data for the driver identification task. In terms of car following patterns, the paper suggests to use the distance to the following car and velocity, map them to a 2D space, and then use a Gaussian Mixture Model (GMM) to obtain an optimal velocity model specific to each driver. Pedal operation patterns were also modeled using another GMM. Instead of using the raw pedal operations patterns, the authors used spectral features derived from the raw data to better identify driver-specific characteristics. The paper had used two datasets for experimentation, one based on simulation data (for 12 drivers) and the other based on real-world data collected for 276 drivers using one vehicle for all drivers and based on a data collection frequency of 100Hz. For simulation dataset, their best accuracy reached 72.9\% using gas pedal operation data, distance to the following car, and velocity with respect to distance to the following car. For the real-world dataset, they achieved an accuracy of 76.8\% by using gas and brake pedal operation data only. 
Enev et al. \cite{enev2016automobile} proposed an ensemble-based solution for driver identification that relies on CAN-bus data, collected for 15 drivers using one vehicle, where each driver drove through the same path on a highway and performed a set of maneuvers in a parking lot. The ensemble classifier model relies on several off-the-self classifiers including Support Vector Machine, Random Forest, Naive Bayes, and K-Nearest-Neighbor (KNN). Using 16 attributes collected using CAN-bus, they were able to achieve an accuracy of 100\%. 
Using one vehicle to collect telemetry data, although provides a fair and sound dataset for modeling and prediction, is not applicable in real-world when building models for a large group of drivers.  

Using acceleration and deceleration events extracted from naturalistic driving behavior for driver identification was proposed by Fung et al. \cite{fung2017driver}. Extracted events from driving data were characterized using five groups of attributes that are duration, speed, acceleration, jerk, and curvature (13 attributes in total). The authors used a dataset of 230K hours of driving data collected from 14 elderly drivers, which comprised 17K acceleration and 17K deceleration events. An accuracy of 50\% was achieved to classify test trajectories taken from 14 drivers by employing a linear discriminative analysis (LDA) model. 
Similarly, Li et al. \cite{li2018driver} proposed using accelerometer data for driver identification. Their proposal relies on statistical features (such as average, standard deviation, quartiles, kurtosis, and skewness) derived from raw accelerometer observation timeseries. They used several off-the-shelf classifiers for modeling, which are KNN, Random Forest, Multi-layer Perceptron (MLP), and AdaBoost, as well as an ensemble of these models. The dataset used in this research comprised 10 taxi drivers, where 16 hours of driving data were collected for each, using calibrated accelerometer sensors that collected data with a frequency of 100Hz. The best performance achieved using the Random Forest model which was about 70\%. 
Relying on small-scale datasets, as well as collecting data on a high-frequency basis make these work somewhat impractical in real-world, despite their promising results. 

Driving DNA is introduced by Fugiglando et al. \cite{fugiglando2017characterizing}, where it covers four aspects of driving, namely cautious driving (measured by braking data), attentive driving (measured by turning data), safe driving (measured by speeding data), and fuel efficiency (measured by RPM data). These aspects were measured per driver per trip, and then average of scores per multiple trips represented a driver's score. The idea was tested on a collection of 2,000 trips collected from 53 drivers based on a wide scenario of road types and open traffic conditions. Their analysis showed the uniqueness of behavior of different drivers when represented by the four aspects. However, no quantitative analysis nor results were presented in the paper to show how accurately one can predict a driver's identity from her driving data. 

Dong et al. \cite{dong2016characterizing} presented a deep-learning framework to learn driving style, which used GPS data to encode trajectories in terms of statistical feature matrices. Then, using two deep-neural-network models, one CNN and one RNN, driver identity is predicted for a given trajectory. Using the same type of input, Dong et al. \cite{dong2017autoencoder} proposed an auto-encoder-based neural network model with a shared RNN component to perform representation learning for trajectories, which can be used for driver identification and trajectory clustering. Likewise, Kieu et al. \cite{kieu2018distinguishing} proposed a similar auto-encoder-based architecture, but using CNN as the shared component, and transforming a trajectory to a location-based 3D image as input to perform trajectory clustering and driver identification. Aim of such frameworks is to utilize both spatial and driving behavior information for the task of driver identification, as opposed to our objective which is to only encode driving style information. 

Chen et al. \cite{chen2019driver} proposed an auto-encoder-based deep neural network model for driver identification. First, a negatively-constrained auto-encoder neural network model is used to find the best window size when scanning through the telemetry data. Then, another deep negatively-constrained auto-encoder neural network is employed to extract latent features from telemetry data. To perform classification, some off-the-shelf classifiers were used that are Softmax, Random Tree, and Random Forest. The dataset used in this paper included 10 drivers who drove the same vehicle through the same paths, and 51 CAN-bus-based features were collected with a frequency of 1Hz. Their best accuracy based on this dataset is reported as 99\%. 
El Mekki et al. \cite{el2019improving} introduced a deep neural network model for driver identification based on telemetry data. The proposed model has two parallel components, one CNN and the other a long short term memory (LSTM) neural network, such that their output is combined using a sigmoid layer. By using four datasets of different sizes and characteristics, comprehensive experiments were performed. Their best accuracy obtained for different datasets are 89.9\% (CAN-bus data collected for 10 drivers using one vehicle), 95.1\% (smartphone data collected for 6 drivers on different vehicles who drove on two different roads), 62.1\% (several sensory data collected for 10 drivers on the same road), and 93.9\% (simulation data for 68 drivers who drove through the same road based on several distraction-based conditions). 
Given the constraints to build datasets used in these work and their limited sizes, one may not be able to replicate their results in real-world. Besides, these promisingly good results might be impacted by spatial memorization as we discuss in this paper. 

Driver prediction based on information collected for a {\em turn} segment is proposed by Hallac et al. \cite{hallac2016driver}, using a hand-crafted rule-based classifier, which at most classifies a group of 5 drivers. Driver prediction for a group of 30 drivers using a Random Forest model, as proposed by Wang et al. \cite{wang2017driver}, achieved an accuracy of $100\%$ for predicting the actual driver, where such result might be due to using high spatially similar trajectories taken from a small group of drivers. Data in both \cite{hallac2016driver} and \cite{wang2017driver} came from the CAN-bus. 
Using equipped vehicles with a variety of sensors and cameras installed inside and outside the vehicle, as well as information coming from the CAN-bus, Ezzini et al. \cite{ezzini2018behind} utilized a set of tree-based models for driver prediction. In \cite{chowdhury2018investigations}, the authors proposed a driver prediction model only by using GPS data. They used GPS data to create a statistical feature vector of size 138 for each trajectory. Then, using a Random Forest classifier they classified small groups of 4 to 5 drivers \cite{chowdhury2018investigations}. 
Using a large set of input features coming from CAN-bus communication which includes 665 features, Hallac et al. \cite{hallac2018drive2vec} proposed a deep-learning model to build embedding for input trajectory data. Then, they used the embedding for a variety of tasks, including driver prediction using a regression model. 

Unlike several of existing work (such as \cite{enev2016automobile,hallac2016driver,wang2017driver,fugiglando2017characterizing,fung2017driver,li2018driver,ezzini2018behind,chowdhury2018investigations,hallac2018drive2vec,chen2019driver,el2019improving}), we model and predict based on a large set of drivers and trajectories. Also, for the first time in literature, we do trajectory filtering based on the spatial similarity between input trajectories for each driver, to prevent spatial memorization and perform driver identification solely based on driving style representation learning. Moreover, we specifically design a deep-neural-network architecture to better capture the driving style for the task of driver identification. 
Lastly, unlike several studies that collected sensory data with high frequencies (e.g., \cite{miyajima2007driver,li2018driver}), used the same vehicle to collect data for all driver (e.g., \cite{miyajima2007driver,enev2016automobile,chen2019driver,el2019improving}), or used precisely instrumented vehicles to collect telemetry data (e.g., \cite{li2018driver,ezzini2018behind,hallac2018drive2vec}), our proposal relies on data collected from a large group of drivers who use their own vehicles, and data were collected based on a reasonable data collection frequency (i.e., 1Hz), which is completely scalable in real-world.

\section{Proposed Model}
\label{sec:architecture}
In this section, we first describe the feature encoding process. Then, we describe our deep-neural-network architecture to learn driving style, called Deep Convolutional Recurrent Neural Network or D-CRNN for short. 

\subsection{Feature Encoding}
\label{sec:feature}
\begin{figure}
    \centering
    \includegraphics[scale=0.45]{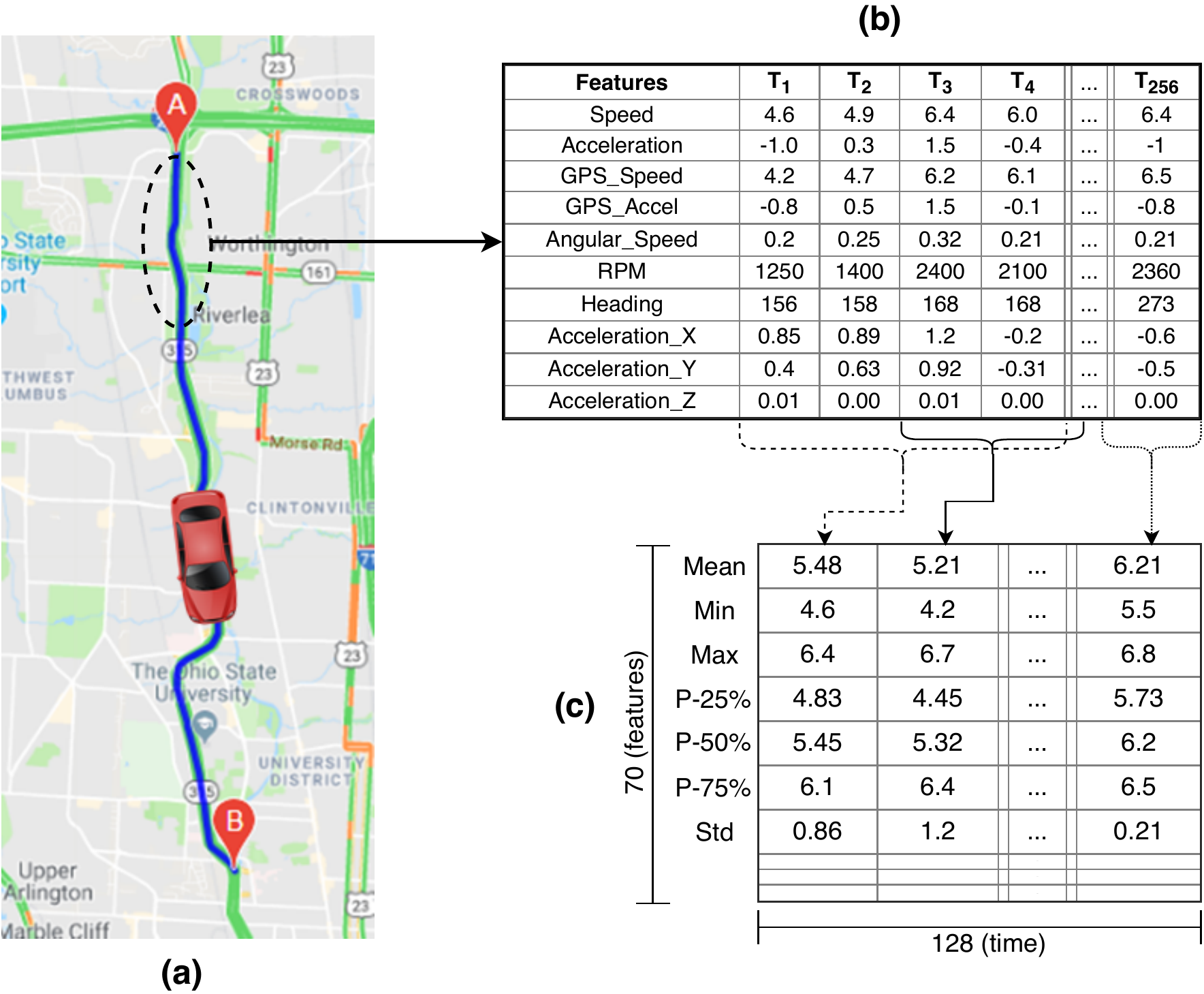}
    \caption{Feature Encoding Process: from (a) raw trajectory, to (b) basic feature map, and to (c) aggregate feature map. In this example we have $L_1 = 256$ and $L_2 = 4$.}
    \label{fig:feature_enc}
\end{figure}

Feature encoding is the process of transforming a trajectory into the form of an input for a machine-learning model. To do so, we adapt a method applied in \cite{dong2016characterizing,dong2017autoencoder}, which is illustrated by Figure~\ref{fig:feature_enc} and its main steps are described below. 

\vspace{5pt}
\noindent \textbf{\normalsize Fixed-Length Segments.} Given a trajectory $T = \langle p_1, p_2, \dots, p_n\rangle$, we first partition that to smaller sub-trajectories (or segments) by a window of size $L_1$, with a shift of $L_1/2$. The overlap between neighboring segments helps to prevent information loss during the partitioning. We choose $L_1 = 256$. 

\vspace{5pt}
\noindent \textbf{\normalsize Basic Feature Map.} After partitioning $T$ to a set of segments $S_T = \{s_1, s_2, \dots, s_k\}$, where $k=2n/L_1-1$, the next step is to generate a feature map (matrix) for each segment $s \in S_T$ which encodes several attributes for each data point. Potentially, we use the following attributes to describe each data point: (1) Speed, (2) Acceleration, (3) GPS\_Speed, (4) GPS\_Acceleration, (5) Angular\_Speed, (6) RPM, (7) Head, (8) AclX, (9) AclY, and (10) AclZ. We call these {\em basic features}. 
Assuming that we select a subset $\mathcal{F}$ of basic features, for a segment of length $L_1$, we represent that by a basic feature map of size $|\mathcal{F}| \times L_1$ (see Figure~\ref{fig:feature_enc}--(b) as an example). 

\vspace{5pt}
\noindent \textbf{\normalsize Aggregate Feature Map.} The last step is to transform a basic feature map to an {\em aggregate} feature map which encodes more high level driving data instead of point-wise, low-level data, and also deals with outliers. In this way, we put each $L_2$ ($L_2 < L_1$) columns of a basic feature map into a frame with shift of $L_2/2$, and for each frame we calculate seven statistics {\em mean}, {\em minimum}, {\em maximum}, $25^{th}$, $50^{th}$, $75^{th}$ {\em percentiles}, and {\em standard deviation}. Describing each data point with a subset $\mathcal{F}$ of basic features, the result of this process is a matrix of size $7|\mathcal{F}| \times 2L_1/L_2$. We choose $L_2 = 4$. See an example in Figure~\ref{fig:feature_enc}--(c). 

\subsection{D-CRNN}
\begin{figure}[ht]
    \centering
    \includegraphics[scale=0.42]{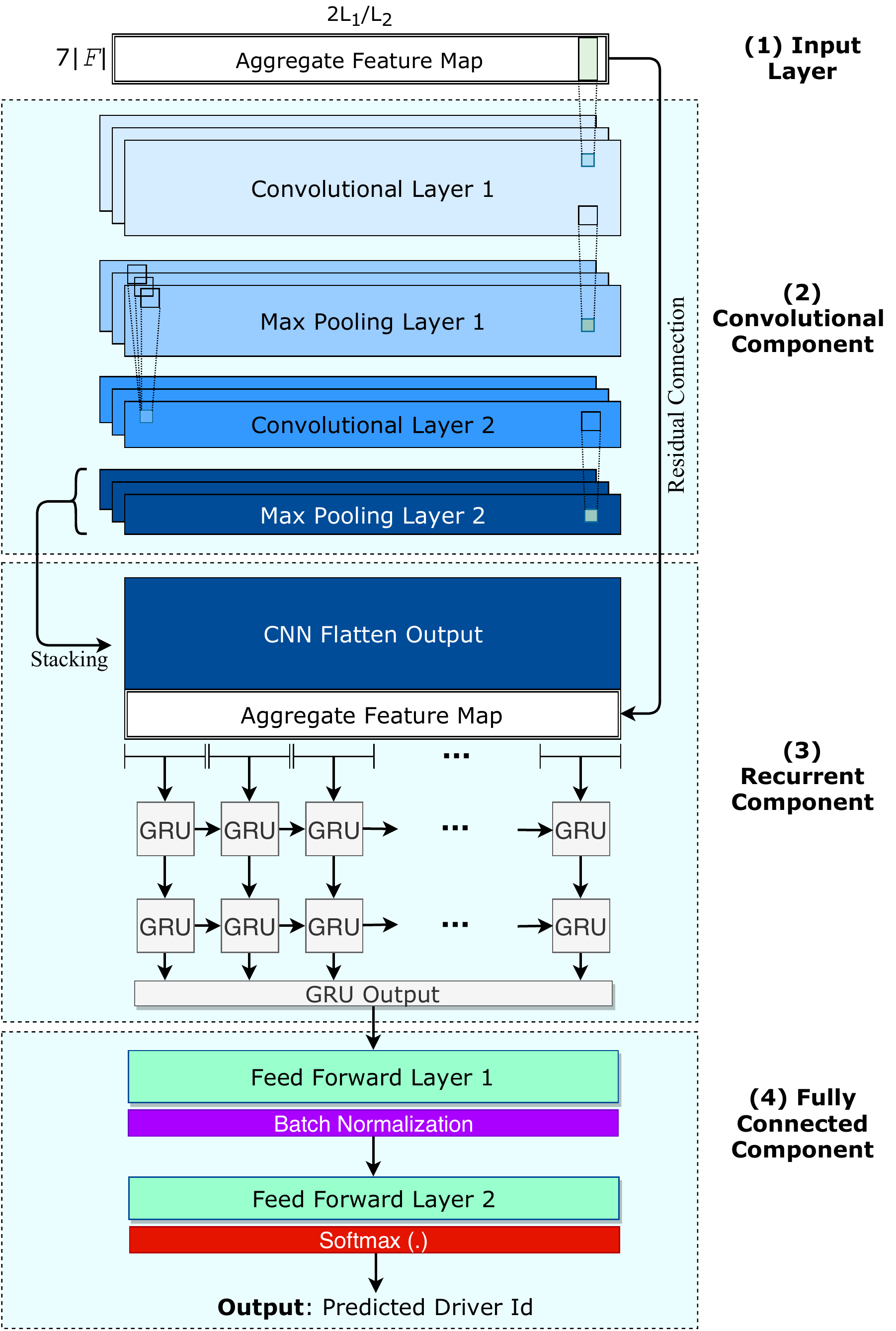}
    \caption{D-CRNN architecture overview: the model consists of four components: 1) Input Layer, 2) Convolutional (CNN), 3) Recurrent (RNN), and 4) Fully Connected (FC) component.}
    \label{fig:dcrnn}
\end{figure}

Deep Convolutional Recurrent Neural Network -- or D-CRNN -- comprises CNN and RNN models, which aims to derive effective representations of driving style. The idea is to use CNN to extract semantic driving patterns (e.g., a turn) from the input trajectory, and RNN to leverage sequential properties to encode dependencies between the patterns. Figure~\ref{fig:dcrnn} illustrates the overall D-CRNN architecture, and we describe its major components as follows. 

\begin{itemize}[leftmargin=*]
    \item \textbf{\normalsize Input Layer}: The layer utilizes the aggregate feature map that is described in Section~\ref{sec:feature}. This feature map is of size $7|\mathcal{F}|\times \big(2L_1/L_2\big)$, where $\mathcal{F}$ is a subset of basic features, and $2L_1/L_2$ is size of the time-axis.
    
    \item \textbf{\normalsize Convolutional Component}: This component seeks to extract semantic driving patterns (e.g., a turn, braking event, etc.). Regarding the input which is represented as an image (or matrix), we believe that these semantic patterns could be characterized as distinguishing visual patterns. Hence, a convolutional network is an appropriate choice to extract such patterns. This component includes two convolutional layers, each followed by a max-pooling to downsample, and dropout (see \cite{srivastava2014dropout}) on top of the pooling layer. The feature maps generated by the second pooling layer are stacked and used as input for the recurrent component. The first convolutional layer has 16 filters, each with a kernel of size $7|\mathcal{F}| \times 5$, which means convolution only on time axis. The second convolutional layer also has 16 filters, each with a kernel of size $3 \times 3$, and both convolutional layers use stride of size $1$. The pooling operations in both layers are performed on feature axis, each with pool size $8\times 1$ and stride $1$. By using zero-padding, we maintain the size of the time-axis the same across convolutional and pooling layers. Thus, the output of the second max-pooling layer after the stacking is of size $16|\mathcal{F}'| \times \big(2L_1/L_2\big)$, where $\mathcal{F}'$ is size of the feature axis after the last max-pooling, and $|\mathcal{F}'| < |\mathcal{F}|$. The activation function for both convolutional layers is the rectified linear unit (or ReLU), and we use a dropout probability of $50\%$ after each pooling layer.
    
    \item \textbf{\normalsize Recurrent Component}: This component aims to leverage the sequential properties of the input over time to better encode dependencies between driving patterns. The stacked output of the convolutional component is concatenated with the aggregate feature map (i.e., the input of the model) along their feature axis, to make a {\em residual connection} (see \cite{he2016deep}), which is used by recurrent component as input. In this way, the input for the recurrent component is of size $\big(16|\mathcal{F}'| + 7|\mathcal{F}|\big) \times \big(2L_1/L_2\big)$. In the recurrent component, we have two layers of Gated Recurrent Unit (GRU) cells, stacked on each other. Each layer has $100$ GRU cells with a dropout probability of $50\%$. Through our hyper-parameter tuning, we found GRU cells to provide the best results in comparison to the vanilla RNN and Long Short Term Memory (LSTM) cells.  
    
    \item \textbf{\normalsize Fully Connected Component}: The output of the second GRU layer is used as input to the two fully connected feed-forward layers, where we have batch normalization \cite{ioffe2015batch} and dropout (with probability $50\%$) after the first fully connected layer. The first layer has $100$ hidden neurons and uses $sigmoid$ as its activation function. The output of the second fully connected layer provides probability values for different class labels (i.e., driver ids) and a softmax is used to make the final prediction. 
\end{itemize}

The usage of dropout is to prevent overfitting \cite{srivastava2014dropout}, especially when we have a deep network, which has proven to be effective in many deep learning applications \cite{krizhevsky2012imagenet}. Residual connection helps to better train a deep network \cite{he2016deep}, and we found it beneficial in our application. Assuming that the convolutional component extracts semantic patterns of trajectory, the usage of residual connection provides the context for such patterns and further helps the recurrent component to identify sequential dependencies between the patterns and extract high-level driving style information. Moreover, the usage of batch normalization is to speed up the training process \cite{ioffe2015batch}. Given a set of trajectories $\mathrm{T}$ as training data, we use back-propagation with {\em cross-entropy} loss function (Equation~\ref{eq:ce}) to train the network:

\begin{equation}
    \label{eq:ce}
    \mathcal{L}_{ce} = \sum_{m \in \mathcal{M}}\, \sum_{d \in \mathrm{D}} -\, \mathds{1}(d,m)\,\times\, log\,(prob(\,d\, |\, m\,))
\end{equation}

\noindent Here $\mathcal{M}$ is the set of all feature maps extracted from trajectory set $\mathrm{T}$, $\mathrm{{D}}$ is the set of all drivers whom trajectories $\mathrm{T}$ are taken from, $\mathds{1}(d,m)$ is a binary indicator function that returns 1 if $m$ belongs to $d$ and 0 otherwise, and $prob(\,d\, |\, m\,)$ is the probability of predicting $d$ as driver of $m$, which is the output of our D-CRNN model prior to applying softmax. 

Grid-search is employed to determine the best architecture settings for: 1) optimal number of convolutional, recurrent, and feed-forward layers (choices of 1, 2, and 3 layers), 2) number of filters for convolutional layers (choices of 8, 16, and 32 filters), 3) size of stride for convolutional layers (choices of 1 and 2), 4) size of recurrent and fully connected layers (choices of 50, 100, and 150 neurons), 5) type of recurrent cells (choices of \textit{vanilla RNN}, \textit{GRU}, and \textit{LSTM}), and 6) activation function for fully connected layers (choices of \textit{sigmoid}, \textit{ReLU}, and \textit{tanh}). 

It is worth noting that convolutional recurrent neural network is rather a new concept which has seen notable interests in literature \cite{yue2015beyond,wu2015modeling,sainath2015convolutional,cakir2017convolutional,choi2017convolutional,zuo2015convolutional,wang2017hybrid,liang2019deep}. 
The differences between our proposal and some of the existing models include the usage of residual connection from input to the recurrent component (which helps the recurrent component to simultaneously utilize semantic patterns as well as low-level behavioral data), the usage of batch-normalization, and properties of different layers in each component.  

\section{Dataset}
\label{sec:dataset}
The data used in this research is a large, private dataset provided by an insurance company. Data was collected in Columbus Ohio, from August 2017 to February 2018, using designated devices connected to the OBD-II port of vehicles which decode {\em CAN-bus} data. Additionally, each device encapsulates several sensors, including {\em gps}, {\em accelerometer}, and {\em magnetometer}. The data collection rate is one record per second (i.e., 1Hz). In total, we have about 4,500 drivers and 836K trajectories in our dataset. Table~\ref{tab:dataset} provides some additional details about the data, including $50^{th}$ percentile (P50) on the duration and traveled distance of each trajectory. Before using the data, we performed several preprocessing, filtering, and sampling steps which we describe them next. Also, we elaborate on why we require each step. 

\begin{table}[ht!]
    \small
    \centering
    \caption{Details on trajectory dataset, collected from Aug 2017 to Feb 2018 in Columbus Ohio. P50 is $50^{th}$ percentile.}
    \setlength\tabcolsep{2pt} 
    \begin{tabular}{| c | c | c | c | c | c |}
        \hline
        \rowcolor{Gray}
        \# Drivers & \# Trajectories & \begin{tabular}{@{}c@{}} Total Travel\\ Time\end{tabular} & \begin{tabular}{@{}c@{}} Total Travel\\ Distance\end{tabular} & \begin{tabular}{@{}c@{}} P50 of \\ Duration\end{tabular} & \begin{tabular}{@{}c@{}} P50 of \\ Distance\end{tabular} \\
        \hline
        4,476 & 835,995 & 221,895 hours& 11,174,400 km & 11 min & 6 km\\
        \hline
    \end{tabular}
    \label{tab:dataset}
\end{table}

\subsection{Data Preprocessing}
Several data preprocessing steps are performed to ensure the quality of data. To prevent memorization of origin and destination, we removed the first and the last two minutes of each trajectory. Additionally, we set the minimum duration of a trajectory to be 10 minutes and the maximum to 30 minutes. This step will not limit the generalizability of our method, but ensures to have enough data for each trajectory and helps to achieve reasonable running time by avoiding long trajectories\footnote{In our dataset, the $90^{th}$ percentile of trajectory duration is about 30 minutes.}. Lastly, we remove those trajectories with any missing attribute to preserve consistency in data.

\subsection{Data Filtering and Sampling}
\label{subsec:data_filtering}
One important step which is overlooked in literature is to filter input trajectories based on {\em spatial similarity}. We define the similarity between two trajectories $T_1$ and $T_2$ in terms of a spatial, point-wise matching function. We call two points $p_i$ and $p_j$ ($p_i \in T_1$ and $p_j \in T_2$) to be matched, if their haversine distance \cite{haversine} is lower than a predefined threshold $\tau$. To find similarity between two trajectories $T_1$ and $T_2$, we propose Algorithm~1. In this algorithm, $p_{1i}$ and $p_{2j}$ are two data points taken from trajectories $T_1$ and $T_2$, respectively, and we use distance threshold $\tau = 100$ (meters). Based on this algorithm, a data point from one of the trajectories can be matched with at most one data point from the other trajectory. Note that similarity measure, as defined here, is a spatial concept that discloses the spatial similarity of two trajectories. Having a dataset with (potential) high spatial similar trajectories for each driver is not an appropriate source of data to learn driving style; because the models can easily learn to memorize geolocation data. Instead, we perform similarity-aware sampling to create our train and test sets. Here we employ three sampling strategies, {\em Threshold-based}, {\em Stratified}, and {\em Random} sampling. 

\begin{algorithm}[ht]
\caption{Spatial Similarity Scoring}
\DontPrintSemicolon
\KwIn{Trajectories $T_1$ and $T_2$, Distance threshold $\tau$}
$matched\_set = []$\;
\For {$i=1$  \textbf{to}  $|T_1|$}
{
    \For{$j=1$  \textbf{to}  $|T_2|$}  
    {
        $dist = haversine(p_{1i}, p_{2j})$\;
        \If {$\big(dist < \tau\big)$ \textbf{and} $\big(p_{2j} \notin matched\_set\big)$}
        {
            $matched\_set.append(p_{2j})$\;
            $break$\;
        }
    }
}
$score = |matched\_set|/min(|T_1|,|T_2|)$\;
\KwOut{score}
\label{algo:spatial_similarity}
\end{algorithm}

\subsubsection{Threshold-Based Sampling}
The idea is to create several subsets of the same size but using different {\em similarity thresholds} to sample trajectories per drivers. We use Algorithm~\ref{alg:threshold_based} for this purpose. By the choice of maximum similarity threshold ($\nu$) from the set $\{0.2, 0.25, 0.3\}$, we generate three sample sets, where the one which is produced by the lowest threshold (i.e., $0.2$) is the most {\em strict} set. As shown by Table~\ref{tab:data_describe}, the $90^{th}$ percentile (i.e., $P90$) on the pairwise similarity between trajectories taken from a driver significantly increases as we increase the similarity threshold.

\begin{algorithm}[ht]
    \caption{Threshold-based Trajectory Sampling}
    \begin{algorithmic}[1]
        \STATE \textbf{Input}: a dataset of drivers and trajectories $(\mathrm{D},\mathrm{T})$, and similarity threshold $\nu$. 
        \vspace{5pt}
        \STATE For each driver $d \in \mathrm{D}$, calculate similarity score between each pair of trajectories taken from $d$ using Algorithm 1. 
        \vspace{5pt}
        \STATE Sample a subset of trajectories for each driver $d \in \mathrm{D}$, which in that the maximum similarity between any pair of trajectories is lower than threshold $\nu$. Call this set $(\mathrm{D}', \mathrm{T}')$, where $\mathrm{D}' \subset \mathrm{D}$ and $\mathrm{T}' \subset \mathrm{T}$. 
        \vspace{5pt}
        \STATE \textbf{Output}: randomly select $50$ drivers in set $\mathrm{D}'$ which for them we have at least $50$ trajectories in set $\mathrm{T}'$. 
    \end{algorithmic}
    \label{alg:threshold_based}
\end{algorithm}

\subsubsection{Stratified Sampling}
A slightly different idea is to create sets with a larger number of drivers while keeping the distribution of pairwise similarity between trajectories the same. Here we employ Algorithm~\ref{alg:stratified_based}, to first create different buckets of trajectories for a driver (based on pairwise similarity), and then uniformly sample from each bucket. In terms of input, we set $N=50$, draw $M$ from set $\{50, 100, 150, 200\}$, and use threshold set $\{0.2, 0.25, 0.3\}$. Therefore, we generate four different sets as shown in Table~\ref{tab:data_describe}. As one could expect, the $90^{th}$ percentile of pairwise trajectory similarity values are almost the same across different sample sets created by this sampling method. 

\begin{algorithm}[h]
    \caption{Stratified Trajectory Sampling}
    \begin{algorithmic}[1]
        \STATE \textbf{Input}: a dataset of drivers and trajectories $(\mathrm{D},\mathrm{T})$, required number of trajectories $\mathrm{N}$, required number of drivers $\mathrm{M}$, and similarity threshold values $\{\nu_1, \nu_2, \dots, \nu_m\}$. 
        \vspace{5pt}
        \STATE For each driver $d \in \mathrm{D}$, calculate similarity score between each pair of trajectories of driver $d$ using Algorithm 1. 
        \vspace{5pt}
        \STATE For each driver $d \in \mathrm{D}$ and for each trajectory $T$ of this driver, calculate the average pairwise similarity value.
        \vspace{5pt}
        \STATE For each driver $d \in \mathrm{D}$, create $m$ subsets of her trajectories, where the $i^{th}$ subset contains trajectories with their average similarity bounded by interval $[\nu_{i-1}, \nu_i)$. Assume $\nu_0 = 0$. 
        \vspace{5pt}
        \STATE Select a subset $\mathrm{D}'$ of drivers $\mathrm{D}$ that for them we have at least $\frac{\mathrm{N}}{m}$ trajectories in each subset. 
        \vspace{5pt}
        \STATE \textbf{Output}: randomly select $\mathrm{M}$ drivers from $\mathrm{D}'$, and for each driver randomly select $\frac{\mathrm{N}}{m}$ trajectories from each subset.
    \end{algorithmic}
    \label{alg:stratified_based}
\end{algorithm}

\subsubsection{Random Sampling}
To show the impact of high spatial similarity in trajectory data for the task of driver identification, we create several random sample sets. Here, using the original dataset, we first sample four groups of drivers of size 50, 100, 150, and 200. Then, for each group, we randomly sample 50 trajectories for each driver. The $90^{th}$ percentile of pairwise trajectory similarity for these sets is assumed to be the same as the original data which is about $0.607$. 

In total, we created 11 sample sets; notation and a short description for each set are provided in Table~\ref{tab:data_describe}. 

\begin{table}[h!]
    \centering
    \caption{Sampled trajectory datasets, using {\em Threshold-based}, {\em Stratified}, and {\em Random} sampling strategies. P90 is the $90^{th}$ percentile on pairwise similarity between trajectories.}
    \setlength\tabcolsep{3pt} 
    \begin{tabular}{| c | c | c | c | c |}
        \hline
        \rowcolor{Gray}
        \textbf{Notation} & \textbf{Drivers} & \textbf{\begin{tabular}{@{}c@{}} P90 of \\ Similarity\end{tabular}} &  \textbf{\begin{tabular}{@{}c@{}} Maximum \\ Similarity\end{tabular}} & \textbf{Strategy} \\
        \hline
        Tb--50\_0.2 & 50 & 0.116 & 0.2 & Threshold-based \\
        \hline
        Tb--50\_0.25 & 50 & 0.166 & 0.25 & Threshold-based \\
        \hline
        Tb--50\_0.3 & 50 & 0.212 & 0.3 & Threshold-based \\
        \hline
        St--50 & 50 & 0.177 & 0.3 & Stratified \\
        \hline
        St--100 & 100 & 0.171 & 0.3 & Stratified \\
        \hline
        St--150 & 150 & 0.168 & 0.3 & Stratified \\
        \hline
        St--200 & 200 & 0.167 & 0.3 & Stratified \\
        \hline
        Rd--50 & 50 & 0.607 & 1.0 & Random \\
        \hline
        Rd--100 & 100 & 0.607 & 1.0 & Random \\
        \hline
        Rd--150 & 150 & 0.607 & 1.0 & Random \\
        \hline
        Rd--200 & 200 & 0.607 & 1.0 & Random \\
        \hline
    \end{tabular}
    \label{tab:data_describe}
\end{table}

\section{Experiments and Results}
\label{sec:results}
In this section, we present experimental settings and results. We start with a description of baseline models, then describe how to perform trajectory-level prediction. Next, we continue with feature analysis, followed by comparing different models using the best combination of basic features. Finally, we conduct several case studies to further demonstrate the usefulness of our proposal\footnote{The code and sample data are available for reproducibility at \url{https://github.com/sobhan-moosavi/DCRNN}}. 

We use {\em accuracy} as our evaluation metric. Given a set of trajectories $\{T_1, T_2, \dots, T_n\}$, where $T_i$ is taken from driver $d_i$, and $prediction(.)$ is a function which predicts identity of driver for $T_i$, we define ``accuracy'' as:
\begin{equation}
  \label{eq:accuracy}  
  Accuracy = \frac{\sum_{i=1}^{n} \mathds{1}_{(prediction(T_i) = d_i)}(T_i)}{n}  
\end{equation}
\noindent where $\mathds{1}$ represents the indicator function.

\subsection{Baselines and Adapted Models}
\label{sec:baseline}
As baseline state-of-the-art models, we use four deep-neural-network-based models and a Gradient Boosting Decision Tree (GBDT) model. 

\subsubsection{CNN Model} 
This is a convolutional neural network model proposed by Dong et al. \cite{dong2016characterizing} for learning driving style. Having the input as aggregate feature map (see Section \ref{sec:feature}), the model employs three convolutional layers on top of the input, each followed by a max-pooling layer. Here, both convolution and pooling operations are over the time axis, hence, both are 1-Dimensional operations. On top of the last max-pooling layer, there are three fully-connected (or dense) layers of size $128$, $128$, and {\em number of drivers}, respectively. 

\subsubsection{RNN Model}
This is a recurrent neural network model proposed by Dong et al. \cite{dong2016characterizing}. The input of the model is the aggregate feature map, where this model utilizes that as a sequential input over time. There are 2 recurrent layers, each with 100 LSTM cells. On top of the output of the second recurrent layer, there are two fully connected layers, where the first layer has 100 neurons and the second layer has as many neurons as the number of drivers which are our labels \footnote{In \cite{dong2016characterizing} authors suggested to use vanilla recurrent units, with ReLU as activation function, and identity matrix to initialize the weight matrix, which provides comparable results to LSTM.}. 

\subsubsection{ARNet Model}
ARNet, proposed by Dong et al. \cite{dong2017autoencoder}, is an auto-encoder based neural network model to build a latent representation for an input trajectory, and also to properly classify a trajectory, where class labels are drivers identities. The model has three parts. The first part includes two GRU layers, each with 100 cells, and one dropout layer. The output of the dropout layer goes to the second and the third part of the model. The second part includes two fully connected layers with 50 and 100 cells, respectively. The aim of the second part is to reconstruct the output of the dropout layer, such that, after the train, the output of the dropout layer can serve as a latent representation for input trajectory. The third part includes a dense layer with as many cells as number of drivers to predict driver identity based on the output of the dropout layer. Similar to RNN Model, ARNet utilizes the aggregate feature map as input. 

\subsubsection{VRAE Model (Adapted Architecture)}
The Variational Recurrent Auto-Encoder (VRAE), proposed by Fabius and Amersfoort \cite{fabius2014variational}, is a generative model, used for learning stochastic latent representations of the input sequences. Being stochastic, the model maps an input to a different latent representation each time. This is a useful property, since it allows the model to associate the input sequence with not just one point, but with nearby points as well. For our task, we treat the raw trajectory data as the sequential input to the VRAE model. However, we train our model to not just be a good generator of trajectories, but learn representations which are highly predictive of the drivers as well. This is done by employing a loss function consisting of two parts - the loss due to likelihood error and the loss due to driver misclassification:
\begin{equation}
    \mathcal{L}_{total} = \lambda\mathcal{L}_{LL} + (1-\lambda)\mathcal{L}_{pred}
\end{equation}
Here, $\mathcal{L}_{LL}$ is the likelihood loss defined in the same way as in \cite{fabius2014variational}, and $\mathcal{L}_{pred}$ is the cross-entropy loss on misclassification of the driver of the trajectory. $\lambda$ is a hyper-parameter which controls the relative importance given to each of the loss terms. Let $y_i$ be the one hot representation of the driver's identity. We define the cross-entropy loss as:
\begin{equation}
    \mathcal{L}_{pred} = \sum_{i=1}^{\text{num\_drivers}}-y_{i}log(\hat{y_i})
    \label{eq:cross_entropy}
\end{equation}
where $\hat{y_i}$, the vector representing model's driver prediction, is calculated as follows:
\begin{equation}
    \hat{y_i} = \text{softmax}(W_{driver}h_{enc} + b_{driver})
\end{equation}
The model employs a GRU encoder to transform the sequential input trajectory into an encoded representation $h_{enc}$. This encoded representation is used for sampling the stochastic latent representation, $z$, as follows:
\begin{align*}
    \mu_z &= W^t_{\mu}h_{enc}+ b_\mu \\
    log(\sigma_z^2) &= W^t_{\sigma}h_{enc} + b_\sigma \\
    z &\sim N(\mu_z, \sigma_z^2)
\end{align*}
The decoder is a GRU as well, whose initial state $h_{dec}$ is generated by applying a linear transformation followed by $tanh$ activation on the latent representation $z$:
\begin{equation}
    h_{dec} = tanh(W_{dec}z + b_{dec})
\end{equation}
The input for VRAE model is also the aggregate feature map.

\vspace{-3pt}
\subsubsection{GBDT Model}
This model is known to be effective for the task of driver identification \cite{kaggle2015driver}. Here we use the set of hand-crafted features as input, described in \cite{dong2016characterizing}. This set includes 321 features. We also modified the original set of features by using the following basic features: Speed, Acceleration, Acceleration\_Change, RPM, RPM\_Change, and Angular\_Speed. By using these features, we represent a trajectory with a vector of size 384. The model based on the original hand-crafted features is termed {\em GBDT-original}, and the one based on the modified features is termed {\em GBDT-modified}. 

\subsection{Trajectory Level Prediction}
All models, except GBDT models, perform segment-level prediction. To obtain trajectory-level prediction, we first obtain the {\em probability vector} for each segment. The probability vector for segment $j$ of trajectory $T$ can be represented as $\langle \rho_{j1}, \rho_{j2}, \dots, \rho_{jm}\rangle$, where $m$ is the number of drivers in a set (i.e., potential labels). This vector is the output of a model before the last softmax layer. Then, for a trajectory $T$ which has $k$ segments, we create a trajectory-level average probability vector as follows:
\begin{equation}
    Prob\_Vector(T) = \frac{1}{k} \sum_{j=1}^{k}\langle\rho_{j1}, \rho_{j2}, \dots, \rho_{jm}\rangle
    \label{eq:avg_prob}
\end{equation}
Then, $\textit{softmax}(Prob\_Vector(T))$ provides the trajectory-level prediction. Note that in our experiments we only report trajectory-level accuracy, and omit the segment-level results in favor of space. 

\subsection{Feature Analysis}
\label{subsec:res_feat_analysis}

This experiment explores the contribution of the basic features, extracted from trajectory data, to the learning of driving style. In total, we have 10 basic features to describe a data point (see Section~\ref{sec:feature}). We use the proposed D-CRNN model in this experiment. For training, we use mini-batches of size 256; 85\% of trajectories taken from each driver in train and 15\% for test; and Root Mean Square Propagation optimizer\cite{tieleman2012divide} (RMSProp for short) with initial learning rate $5e-5$, momentum $0.9$, and epsilon $1e-6$ as optimizer function. Here we compare different combinations of basic features based on {\em Accuracy} (see Equation~\ref{eq:accuracy}). 
In terms of data, we used data sample sets Tb--50\_0.2, St--200, Rd--50, and Rd-200. Table~\ref{tab:feature_analysis} shows the results of this experiment, using 13 different subsets of basic features, including the entire set, and we summarize our observations as follows. 

\begin{itemize}[leftmargin=10pt]
    \item {\em Best Subset}: The best combination of features is found to be ``Speed'', ``Acceleration``, and ``RPM'', across all four datasets. Even using the entire set of basic features did not result in a better accuracy. Based on this observation, we argue that having these three features is enough to learn the driving style of an individual for use in driver identification. 

    \item {\em Most Effective Feature}: The most effective feature is found to be RPM. While this observation may be related to the uniqueness of RPM observations for different vehicle brands, we reject this hypothesis for two reasons. First, using the set ``Tb--50\_0.2'' we have the least similarity among trajectories of a driver, which means that it is very unlikely to observe the same series of RPM values for different trajectories. Second, in set ``St--200'' we used a much larger set of vehicles to decrease the possibility of having vehicles with unique characteristics. {\it We also note that RPM is the outcome of an essential driving behavior (i.e., pressing the gas pedal), therefore we expect it to be a robust predictor of driver identity}. 

    \item {\em GPS Related Features}: This experiment revealed a significant difference between Speed and Acceleration generated based on GPS coordinates, versus the ones reported directly by the vehicle using CAN-bus. Although GPS is a reliable source to obtain velocity information, the direct usage of coordinates to obtain speed and acceleration may not be the best practice. Instead, in practice, the Doppler shift of the received carrier frequencies is used to determine the velocity of a moving receiver. It is also known that Doppler-derived velocity is far more accurate than velocity obtained based on differencing GPS coordinates\footnote{For more details please visit \url{https://rb.gy/u0rlmo}.}. Nevertheless, in the absence of Doppler shift data, we can only leverage position estimations to estimate velocity, which does not result in the best representation of driving style and driver identification. 
    

    \item {\em Memorization versus Learning Driving Style}: The low prediction accuracy for a feature like ``Head'' is an indicator of the sampling quality to avoid memorization. If prediction accuracy based on ``heading`` was high, as it is when using random sample sets Rd--50 and Rd-200, this would have potentially indicated that the model is doing spatial memorization instead of learning driving style. Since for a given driver, their commute patterns can be represented by a limited number of sequences of heading (or bearing) observations\footnote{Although heading (or bearing) does not directly encode any location information, a sequence of heading observations can be attributed to a specific route. Thus, it is not difficult to use a sequential neural network model to find similarities between such sequences and predict driver identity.}. Likewise, a comparison between results obtained using random sets and other sample sets further validates our initial hypothesis on the significant impact of spatial information. 
\end{itemize}

\noindent We also note that the low data collection frequency (i.e., 1Hz) could be a potential reason for relatively weak prediction results based on accelerometer data. 

\begin{table*}[ht!]
    \centering
    \caption{Driver prediction based on different combinations of basic features using D-CRNN, tested on data sample sets {\em Tb--50\_0.2}, {\em Rd--50}, {\em St--200}, and {\em Rd--200}. Results are reported based on Accuracy metric (see Equation~\ref{eq:accuracy}). RPM is the outcome of an essential driving behavior, therefore we expect it to be a robust predictor of driver identity. }
    \label{tab:feature_analysis}
    \setlength\tabcolsep{10pt} 
    \begin{tabular}{cc c c c c}
        \toprule
        \textbf{Basic Features} & & \textbf{Tb--50\_0.2} & \textbf{Rd--50} &  \textbf{St--200} &  \textbf{Rd--200}\\
        &  &  &  & \\
        Entire Feature Set &&  49.07\% &  70.96\% &  46.27\% &   61.77\%  \\
        RPM && 54.75\% &   75.00\% &   54.37\% &   65.34\%  \\
        Head && 2.49\% &  23.66\% &  1.00\% & 14.84\%  \\
        Speed && 20.90\% &  34.92\% & 16.68\% & 31.62\%  \\
        RPM and Head && 45.61\% & 70.89\% & 47.37\% & 62.02\%  \\
        RPM, Head, AclX, AclY, and AclZ && 40.38\% & 66.84\% & 51.17\% & 61.83\%  \\
        GPS\_Speed, GPS\_Accel, and Angular\_Speed && 8.66\% & 15.86\% & 4.08\% & 12.26\%  \\
        GPS\_Speed, GPS\_Accel, and RPM && 45.57\% & 71.40\% & 48.28\% & 61.02\%  \\
        Speed, Accel, and Angular\_Speed && 27.27\% & 32.39\% & 17.48\% & 28.24\%  \\
        Speed, Accel, Angular\_Speed, RPM, and Head && 50.00\% & 73.39\% & 52.09\% & 63.81\%  \\
        Speed, Accel, and RPM && \textbf{56.06}\% & \textbf{75.63}\% & \textbf{58.49}\% & \textbf{70.90}\%  \\
        Speed, Accel, AclX, AclY, and AclZ && 27.76\% & 50.00\% & 28.39\% & 43.14\%  \\
        AclX, AclY, and AclZ && 14.78\% & 34.83\% & 22.33\% & 37.52\%  \\
        \toprule
    \end{tabular}
\end{table*}

\subsection{Impact of Spatial Similarity}
\label{subsec:res_data_sim}

This experiment studies the impact of similarity among trajectories, and to assess the ability of different models to learn driving style instead of spatial memorization. Here, we compare different models using following sets: ``Tb--50\_0.2'', ``Tb--50\_0.25'', ``Tb--50\_0.3'', and ``Rd--50''. We use speed, acceleration, and RPM as the basic features. 
For RNN-model, CNN-model, and ARNet we use the same hyper-parameters as suggested in \cite{dong2016characterizing} and \cite{dong2017autoencoder}. For VRAE we use $\lambda = 0.3$, $|h_{enc}| = |h_{dec}| = |z| = 128$ (see Section~2 of Supplementary Material). As before, we use 85\% of trajectories taken from each driver in train and 15\% for the test\footnote{The same train and test sets are used for all the models.}, the size of mini-batches is set to 256, and RMSProp with the same setting as the previous section is used as the optimizer. Table~\ref{tab:res_data_similarity} presents the results of this experiment in terms of average accuracy over three runs. Here, {\em D-CRNN-W/O-BR} shows a variation of D-CRNN without batch-normalization and residual connection. Based on Table \ref{tab:res_data_similarity}, we make the following observations:

\begin{itemize}[leftmargin=10pt]
    \item {\em Best Model}: The best prediction results are obtained by D-CRNN. In comparison to the state-of-the-art, our proposed architecture achieves about $6\%$ improvement based on the set of trajectories with the least similarity between them (i.e., Tb--50\_0.2). 

    \item {\em Learning versus Memorization}: Increasing the similarity threshold simplifies the driver identification task, as we include more hints in terms of spatial similarity between trajectories. This scenario becomes clearer when using the random sample set, which on-average yields $22\%$ difference in accuracy when compared to the set ``Tb--50\_0.2''. This is an important observation because it justifies why data preprocessing is required to have a sound comparison framework. Also, we claim the consistent trend of better results shows the superiority of our model to better learn the driving style. 

    \item {\em Non-deep Learning Models}: Given the results, we conclude that the modified set of basic feature provides significant improvement in comparison to the case of using the original features. However, the best accuracy obtained by a GBDT model is still lower than the worst accuracy by a deep-neural-network-based architecture (i.e., VRAE model).
    
    \item {\em Results with VRAE}: VRAE generates stochastic latent vectors, which is the reason why it is unable to capture the high-level structure in trajectories as well as other deep learning models. Consequently, the VRAE model suffers in driver prediction.
\end{itemize}
Our experimental results also reveal the importance of batch normalization and residual connection. 

\begin{table}[t]
    \centering
    \caption{Studying the impact of Spatial Similarity on driving style learning and driver identity prediction.}\vspace{-5pt}
    \setlength\tabcolsep{2pt} 
    \begin{tabular}{ c  c  c  c  c}
        \toprule
        \textbf{Model} & \textbf{Tb--50\_0.2} & \textbf{Tb--50\_0.25} & \textbf{Tb--50\_0.3} & \textbf{Rd--50} \\\vspace{-5pt}
        & & & & \\
        GBDT-original \cite{kaggle2015driver} & 5.70\% & 3.09\% & 6.50\% & 25.13\% \\
        GBDT-modified & 25.65\% & 32.87\% & 29.54\% & 49.24\% \\
        CNN-model \cite{dong2016characterizing} & 39.81\% & 44.48\% & 46.88\% & 59.90\% \\
        RNN-model \cite{dong2016characterizing} & 50.69\% & 57.02\% & 58.36\% & 70.73\% \\
        ARNet \cite{dong2017autoencoder} & 56.38\% & 61.22\% & 63.20\% & 75.63\% \\
        VRAE \cite{fabius2014variational} & 26.50\% & 42.20\% & 54.50\% & 63.67\% \\
        D-CRNN & \textbf{59.50}\% & \textbf{64.42}\% & \textbf{66.76}\% & \textbf{78.00}\% \\
        D-CRNN-W/O-BR & 54.40\% & 59.93\% & 63.41\% & 75.63\% \\
        \toprule
    \end{tabular}
    \label{tab:res_data_similarity}
\end{table}

\subsection{Impact of Data Size}
\label{subsec:res_data_size}

Next, we use sample sets of larger sizes, in terms of the number of drivers. Using the same hyper-parameters and setting as the last section, the results of this experiment are shown in Table~\ref{tab:res_data_size}, in terms of accuracy based on three experiments for each model and dataset. Note that we used speed, acceleration, and RPM as the basic features. We make the following observations:

\begin{itemize}[leftmargin=10pt]
    \item {\em Best Model}: D-CRNN is the best model when we use more drivers. However, as we increase the number of drivers, our results become closer to RNN-model and ARNet.  
    
    \item {\em Effect of Data Size}: By increasing the number of drivers, we see a decreasing trend of accuracy. However, the slope of drop for some of the models is slower than the others, which demonstrates the ability of those models to better learn driving style, including D-CRNN, ARNet, RNN-model, and CNN-model. 
    
   \item {\em Deep versus Non-deep}: Regarding the results, as we increase the number of drivers, we see more struggle by non-deep methods such as GBDT. However, most of the deep-models successfully handle more drivers and provide noticeably better results. 
\end{itemize}
Note that comparing the results of random samples with the sets with uniform similarity distributions, once again, reveals the impact of spatial similarity that leads to significant overestimation of the capability of different models to learn driving style. 

\begin{table*}[h]
    \centering
    \setlength\tabcolsep{10pt} 
    \caption{Studying the effect of Data Size on driving style learning and driver identity prediction.}
    \label{tab:res_data_size}
    \begin{tabular}{ c  c | c || c | c || c | c || c | c }
        \toprule
        \textbf{\small Model} & \small \textbf{St--50} & \small \textbf{Rd--50} & \small \textbf{St--100} & \small
        \textbf{Rd--100} & \small \textbf{St--150} & \small \textbf{Rd--150} & \small \textbf{St--200} & \small \textbf{Rd--200} \\\vspace{-8pt}
        &  &  &  &  &  &  & &\\
        GBDT-original \cite{kaggle2015driver} & 6.58\% & 25.13\% & 2.23\% & 23.24\% & 2.61\% & 20.67\% & 2.18\% & 19.53\% \\
        GBDT-modified & 30.26\% & 49.24\% & 20.18\% & 36.62\% & 16.28\% & 31.84\% & 13.37\% & 29.81\% \\
        CNN-model \cite{dong2016characterizing} & 42.37\% & 59.90\% & 44.17\% & 52.66\% & 39.14\% & 53.27\% & 40.27\% & 53.49\% \\
        RNN-model \cite{dong2016characterizing} & 58.07\% & 70.73\% & 57.23\% & 67.48\% & 55.29\% & 68.48\% & 57.34\% & 67.32\% \\
        ARNet \cite{dong2017autoencoder} & 62.33\% & 75.12\% & 59.47\% & 71.08\% & 56.18\% & 70.05\% & 57.41\% & 67.25\% \\
        VRAE \cite{fabius2014variational} & 48.58\% & 63.67\% & 9.33\% & 44.80\% & 8.00\% & 11.50\% & 3.00\% & 11.00\% \\
        D-CRNN & \textbf{65.35}\% & \textbf{78.00}\% & \textbf{61.64}\% & \textbf{73.56}\% & \textbf{56.92}\% & \textbf{71.89}\% & \textbf{57.50}\% & \textbf{69.66}\% \\
        D-CRNN-W/O-BR & 61.58\% & 75.63\% & 57.97\% & 68.42\% & 54.13\% & 66.75\% & 54.51\% & 68.15\% \\
        \toprule
    \end{tabular}
    \vspace{10pt}
\end{table*}

\subsection{Case Study I: A Real-World Application}
\label{subsec:realworld}

\begin{table*}[h]
    \centering
    \caption{Studying the effect of training on threshold-based sample set and testing on random sets.}\vspace{-5pt}
    \setlength\tabcolsep{10pt} 
    \begin{tabular}{ c  c  c  c }
        \toprule
        \textbf{Model} & \begin{tabular}{@{}c@{}} \textbf{Tb--50\_0.2} \\ (Random test set)\end{tabular} & \begin{tabular}{@{}c@{}} \textbf{Tb--50\_0.25} \\ (Random test set)\end{tabular} & \begin{tabular}{@{}c@{}} \textbf{Tb--50\_0.3} \\ (Random test set)\end{tabular} \vspace{4pt}\\\vspace{-10pt}
        & & & \\
        GBDT-original \cite{kaggle2015driver} & 6.25\% & 8.75\% & 10.00\% \\
        GBDT-modified & 28.25\% & 32.50\% & 36.00\% \\
        CNN-model \cite{dong2016characterizing} & 43.92\% & 52.00\% & 54.13\% \\
        RNN-model \cite{dong2016characterizing} & 59.75\% & 60.67\% & 64.33\% \\
        ARNet \cite{dong2017autoencoder} & 64.12\% & 67.40\% & 71.25\% \\
        D-CRNN & \textbf{73.25}\% & \textbf{76.17}\% & \textbf{75.75}\% \\
        \toprule
    \end{tabular}
    \label{tab:real_world}
\end{table*}

In a real-world application, we might perform offline pre-processing to filter trajectory data prior to train and apply the model on random test data to identify the identity of drivers in real-time. For this scenario, we modify the datasets used in Section~\ref{subsec:res_data_sim} as follows: given a threshold-based sample set, we keep the train data, but replace the test data with a random sample of trajectories taken from each driver. Again, for each driver, we have 50 trajectories, where $85\%$ of them are used for train and $15\%$ for the test. Using the same models and hyper-parameters, after running each model for three times, the average accuracy results are shown in Table~\ref{tab:real_world}. As one might expect, using random test sets, we obtained significantly better accuracy results (about 25\% on average). 
Such an outcome can be attributed to simplifying the testing condition by increasing the chance of having spatially-similar trajectories in train and test sets. Another interesting observation is the proximity of results when using different threshold-based training sample sets using RNN-model, ARNet, and D-CRNN. This shows the ability of these models to better derive style information instead of memorizing location data, which leads to more stable prediction results. 

\subsection{Case Study II: Driving Style Representation}

\begin{figure*}[t!]
    \centering
    \begin{adjustwidth}{-.5cm}{-.5cm}
        \includegraphics[scale=0.35]{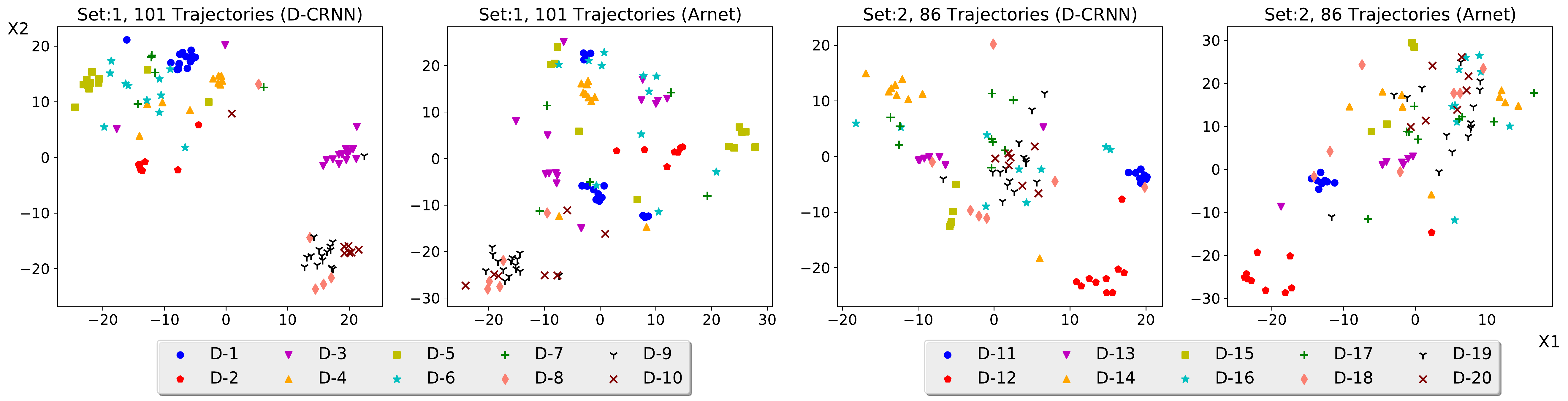}
    \end{adjustwidth}
    \begin{adjustwidth}{-.25cm}{-.25cm}
        \caption{Demonstration of learnt driving style by ARNet\cite{dong2017autoencoder} and D-CRNN, using two random sample sets of 10 drivers. Original latent vectors are 100-D, and we used t-SNE \cite{maaten2008visualizing} for feature selection to represent them in a 2-D space.}
        \label{fig:latent_driving_style}
    \end{adjustwidth}
\end{figure*}

The main objective of this paper is to create models for driver identification by learning driving style. To further assess the accomplishment of this objective, in this section we provide some visualizations using the learned latent representation of trajectories for several drivers. These trajectories are those which we used in the test set. To make the task and evaluation more rigorous, we choose set ``Tb-50\_0.2'' which has the least spatially similar trajectories for each driver. Here we compare two models, D-CRNN and ARNet. In terms of latent representation, we use the output of the first fully connected layer for D-CRNN, and the output of dropout layer for ARNet. As such outputs are for a segment of a trajectory, thus, given a trajectory, we use the {\em average vector} of representations of its segments as trajectory latent representation. For both models, this representation is a 100-dimensional vector. We use t-SNE method \cite{maaten2008visualizing} to perform feature selection and show the latent representations in a 2-dimensional space. 
We randomly sample two sets of 10 drivers. Figure~\ref{fig:latent_driving_style} illustrates the results of this experiment, where the latent representations of trajectories based on D-CRNN and ARNet are shown for both sample sets. Here, the expected outcome is to see trajectories taken from different drivers to be far from each other, while trajectories taken from the same driver to be sufficiently close. Given the results, both methods provide reasonable latent representations, while D-CRNN provides better separation for multiple cases. For example, D-CRNN provides better separation for trajectories taken from drivers 
D-1, D-3, D-5, D-9, D-10, D-12, D-13, and D-14. 

Note that closeness of the trajectories based on their latent representation cannot be a result of spatial similarity, as we use the set of least similar trajectories for each driver. Instead, this is a demonstration of the drivers' fingerprint (or style) which discriminates them from each other.

\subsection{Case Study III: Driver ``Resolution''}
\label{subsec:res_clustering}
Another potential use of driving style information is
for driver ``resolution'' -- analogous to entity resolution. Given a set of trajectories $\mathcal{T}$ taken from $m$ drivers, the task comprises two parts: 1) correctly identifying the number of drivers for an arbitrary set of trajectories $T \subset \mathcal{T}$, sampled from $n$ drivers ($n\leq m$); and 2) correctly partitioning data such that trajectories that belong to a driver $d$ fall into the same partition, while trajectories taken from different drivers fall into the different partitions. The implication of this task is to discriminate across different drivers who share the same vehicle, in order to provide customized insurance policies for individual drivers based on ``how'' they drive. 
Similar to the previous section, we create a latent representation for each trajectory, and use it as input for a clustering algorithm to perform driver resolution. 
As baselines, we use CNN-model, RNN-model, and ARNet. For CNN-model, we use the output of its second fully connected layer after the convolutions as latent representation for a segment of a trajectory. For the RNN-model, we use the output of the first fully connected layer after the recurrent layers as the latent representation. Lastly, we follow the same strategy as the previous section to obtain latent representations based on ARNet and D-CRNN models. 

For this experiment, we use set ``Tb-50\_0.2'', which includes the least spatially similar trajectories for each driver. Using the same input for all four models (i.e., the aggregate feature map based on speed, acceleration, and RPM), the CNN-model and the RNN-model were trained for 200 epochs using the hyper-parameter settings that were recommended in \cite{dong2016characterizing}, the ARNet model was trained for 150 epochs using the hyper-parameter settings that were recommended in \cite{dong2017autoencoder}, and the D-CRNN model was trained based on the same settings that described in Section~\ref{subsec:res_data_sim}. 

We randomly sampled 10,000 subsets of trajectories in ``Tb-50\_0.2'', each included 10 drivers and their trajectories in the test set. Then, we created latent representation for each trajectory based on each trained model, and used them as input for the clustering algorithm. 
Here we report the average estimation error (EE) for the number of clusters and average Adjusted Mutual Information (AMI) score \cite{vinh2010information}. The first metric represents the error in estimating the true number of clusters, where the expectation is that the predicted number of clusters to be close to the number of drivers that trajectories are collected from. The second metric represents the matching between clustering partitions and true labels. 
AMI scores vary from 0 to 1, where an AMI score of 1 represents a perfect match. 

We employed {\it affinity propagation} \cite{dueck2007non} as the clustering algorithm, however, this choice does not limit the generalizability of this experiment. Affinity propagation has two parameters, {\em damping factor} and {\em preference}. We set $\textit{damping factor} = 0.5$ for all four models, and $\textit{preference}=-12$ for CNN-model, $\textit{preference}=-45.5$ for RNN-model, $\textit{preference}=-55$ for ARNet, and $\textit{preference} = -6.5$ for D-CRNN. We empirically found these settings to derive the best results for each model. Table~\ref{tab:clustering_results} shows the results of this experiment. 
Based on the results, our proposal outperforms the baselines by significant margins. With respect to AMI, D-CRNN outperforms the best baseline by about $15\%$, while reduces the error in estimating the true number of clusters. Using the set of least spatially similar trajectories for this task, these results show the usefulness of our proposal in capturing essential driving style information from trajectory data that can be used to determine the number of drivers when given an arbitrary set of trajectories, and better partition trajectories to maximize intra-cluster similarity and inter-cluster dissimilarity. 

\begin{table}[h]
    \centering
    \caption{Driver resolution results based on Adjusted Mutual Information (AMI) \cite{vinh2010information} and Estimation Error (EE). Here we report average and standard deviation (std) based on each metric for both models.}
    \setlength\tabcolsep{2pt} 
    \begin{tabular}{c | c | c | c | c }
        \textbf{Model} & \textbf{Average-AMI} & \textbf{Std-AMI} & \textbf{Average-EE} & \textbf{Std-EE} \\
        \hline
        \hline
        CNN-model \cite{dong2016characterizing} & 0.16 & 0.05 & 1.11 & 0.93 \\
        \hline
        RNN-model \cite{dong2016characterizing} & 0.49 & 0.07 & 1.14 & 0.95 \\
        \hline
        ARNet \cite{dong2017autoencoder} & 0.54 & 0.08 & 1.20 & 0.98 \\
        \hline
        D-CRNN & \textbf{0.62} & 0.07 & \textbf{1.08} & 0.91 \\
    \end{tabular}
    \label{tab:clustering_results}
\end{table}



\section{Conclusion and Future Work}
\label{sec:conclusion}
In this paper, we present a deep-neural-network architecture to learn driving style from trajectory data. The proposed model is inspired by the idea of convolutional recurrent neural networks, to empower the model by extracting semantic patterns from trajectories, and by encoding the sequential dependencies between the patterns. We also propose to sample dissimilar trajectories on a large scale to prevent spatial memorization which could mislead the task and invalidate the results. Our analyses and results demonstrate the superiority of the proposed architecture for driver identification in comparison to the state-of-the-art, based on different test scenarios and datasets. 
We conclude that using deep learning models for the task of learning driver style yields tremendous improvement in comparison to the non-deep learning solutions, especially when using a proper combination of deep-neural-network components. 
Moreover, geolocation bias is an important challenge when we design frameworks based on data that has spatial information, while utilizing such data may not be always an objective. Finally, we note that a subset of data elements coming from CAN-bus is the best combinations to represent the driving style. For future work, we intend to expand the framework for the important task of {\em driver risk prediction}, to be used for usage-based insurance. 

\section*{Acknowledgments}
This work is supported by a grant from the Nationwide Mutual Insurance (GRT00053368), and another from the Ohio Supercomputer Center (PAS0536). Any findings and opinions are those of the authors. 

\bibliography{main.bbl}

\end{document}